\documentclass[conference]{IEEEtran}
\usepackage{times}
\usepackage[numbers,square,comma,sort&compress]{natbib}
\usepackage[utf8]{inputenc} %
\usepackage[T1]{fontenc}    %
\usepackage{url}            %
\usepackage{booktabs}       %
\usepackage{amsfonts}       %
\usepackage{amsmath}
\usepackage{nicefrac}       %
\usepackage{microtype}      %
\usepackage{xcolor}         %
\usepackage{graphicx}       %
\usepackage[font=footnotesize]{caption}
\usepackage{subcaption}     %
\usepackage{algorithm}
\usepackage[noend]{algpseudocode} %
\usepackage{wrapfig}
\usepackage[bottom]{footmisc}
\usepackage{listings}
\usepackage{hyperref}

\allowdisplaybreaks
\title{Training-Time Action Conditioning for Efficient Real-Time Chunking}

\author{%
  Kevin Black \quad%
  Allen Z. Ren \quad%
  Michael Equi \quad%
  Sergey Levine \\
  Physical Intelligence
}

\newcommand{\ba}{\mathbf{a}}
\newcommand{\bv}{\mathbf{v}}

\newcommand{\bo}{\mathbf{o}}

\newcommand{\bI}{\mathbf{I}}
\newcommand{\bA}{\mathbf{A}}

\newcommand{\E}{\mathbb{E}}
\newcommand{\beps}{\boldsymbol{\epsilon}}

\def \PiZeroSix {$\pi_{0.6}$}
\def \PiStarZeroSix {$\pi^*_{0.6}$}

\begin{document}

\maketitle

\begin{abstract}
Real-time chunking (RTC) enables vision-language-action models (VLAs) to generate smooth, reactive robot trajectories by asynchronously predicting action chunks and conditioning on previously committed actions via inference-time inpainting. However, this inpainting method introduces computational overhead that increases inference latency. In this work, we propose a simple alternative: simulating inference delay at training time and conditioning on action prefixes directly, eliminating any inference-time overhead. Our method requires no modifications to the model architecture or robot runtime, and can be implemented with only a few additional lines of code. In simulated experiments, we find that training-time RTC outperforms inference-time RTC at higher inference delays. In real-world experiments on box building and espresso making tasks with the $\pi_{0.6}$ VLA, we demonstrate that training-time RTC maintains both task performance and speed parity with inference-time RTC while being computationally cheaper. Our results suggest that training-time action conditioning is a practical drop-in replacement for inference-time inpainting in real-time robot control.
\end{abstract}

\section{Introduction}
Unlike chatbots or search engines, embodied agents must operate in real time. The feedback loop between an agent's actions and its environment necessitates reactivity --- like a human athlete, an agent cannot simply ``stop and think'' while the outside world changes. However, the ever-increasing size of frontier models makes this more and more difficult. Nowhere is this more evident than in the domain of robot learning, where vision-language-action models (VLAs) consisting of billions of parameters have increasingly been used to control robots at high frequencies to accomplish dexterous tasks. Producing smooth yet reactive trajectories when the model inference latency is in the tens to hundreds of milliseconds is no small challenge.

Real-time chunking (RTC; \citep{black2025real}) presents an approach to this problem that combines action chunking \citep{chi2023diffusion,zhao2024aloha}, flow matching \citep{lipman2022flow}, and inference-time inpainting \citep{song2023pseudoinverse,pokle2023training}. In RTC, action chunks are predicted asynchronously --- the next chunk is generated while the current one is still executing. To ensure continuity between chunks, each generation is conditioned on a frozen \textit{prefix} of previously predicted actions, inpainting the rest. However, the inference-time inpainting method used by RTC introduces additional computational overhead --- and hence latency --- that somewhat defeats the purpose of a real-time execution framework. Empirically, we also find that inference-time inpainting is fundamentally limited in its ability to handle high inference delays.

In this work, we augment RTC with an inpainting method that simulates inference delay at training time and eliminates any inference-time computational overhead. Our method works as a drop-in replacement for inference-time RTC: it requires no modifications to the model architecture or the robot runtime, and can be implemented with only a few additional lines of code. On simulated benchmarks, we find that training-time RTC outperforms inference-time RTC at higher inference delays. In the real world, we demonstrate that training-time RTC can be successfully added by fine-tuning a base model that was \textit{not} pre-trained with action prefix conditioning. By applying training-time RTC to the \PiZeroSix{} VLA \cite{pi06}, we show improved performance over inference-time RTC on two highly complex tasks: box building and espresso making.

\section{Related Work}

\textbf{Action chunking and VLAs.}
Action chunking \citep{zhao2023learning,chi2023diffusion} is the de facto standard in end-to-end imitation learning for visuomotor control. Recently, augmenting vision-language models (VLMs) to produce action chunks has demonstrated great success in robot manipulation, giving rise to vision-language-action models (VLAs)~\citep{rt22023arxiv,collaboration2023open,kim2024openvla,black2024pi,zheng2024tracevla,cheng2024navila,cheang2024gr2generativevideolanguageactionmodel,zhen20243dvla,intelligence2025pi,pertsch2025fast,liu2024rdt}. Subsequently, a plethora of methods have emerged to address the tension between large VLAs and high-frequency control. For example, Gemini Robotics \citep{team2025gemini} and GR00T \citep{bjorck2025gr00t} employ hierarchical VLA designs where the model is split into a heavyweight System 2 (high-level planning) and lightweight System 1 (low-level action generation) component. MiniVLA \citep{belkhale2024minivla} and SmolVLA \citep{shukor2025smolvla} present VLA architectures that are altogether faster and more efficient than most designs, making inference at the edge more feasible. These contributions are orthogonal to ours, and come with their own tradeoffs (e.g., modified network architectures and training recipes).

\textbf{Real-time execution of VLAs.}
The most closely related prior work is real-time chunking (RTC; \citep{black2025real}), which introduces an asynchronous execution framework that serves as a foundation for this work. Also related is SmolVLA \citep{shukor2025smolvla}, which presents an asynchronous execution algorithm that is similar to that of RTC; however, SmolVLA does not solve the inter-chunk discontinuity problem, which leads to out-of-distribution ``jerks'' between chunks. Concurrently to this work, A2C2 \citep{sendai2025leave} and VLASH \citep{tang2025vlash} both solve the discontinuity problem by adding a lightweight correction head and by conditioning on a single future action, respectively. In contrast to VLASH, we condition on a full prefix of future actions.

\section{Preliminaries}

We use the same problem formulation as RTC \citep{black2025real}: we begin with an action chunking policy denoted by $p(\bA_t | \bo_t)$, where $\bA_t = [\ba_{t}, \ba_{t+1}, ..., \ba_{t+H-1}]$ is a chunk of future actions, $\bo_t$ is an observation, and $t$ indicates a controller timestep.
We call $H$ the \textit{prediction horizon}, and at inference time, we roll out each chunk for $s \leq H$ timesteps, where $s$ is the \textit{execution horizon}.

\begin{figure}[tbp]
  \centering
  \includegraphics[width=0.48\textwidth]{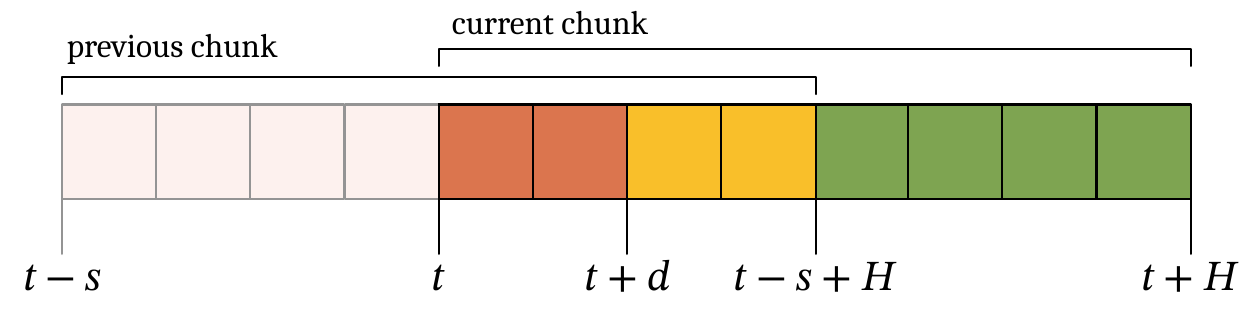}
  \caption{A diagram illustrating two overlapping action chunks. The $d$ actions between $t$ and $t + d$, taken from the previous chunk, are the action prefix (red). From the diagram, we can easily see that we must satisfy the constraint $t + d \leq t - s + H \to d \leq H - s$ to have a valid action prefix. Note that inference-time RTC uses all $H - s$ overlapping actions (red and yellow) to guide the generation of the current chunk, whereas training-time RTC only uses the first $d$ actions (red).}
  \label{fig:diagram}
\end{figure}

To account for model inference, we define the quantity $d$ to be the inference delay in units of controller timesteps. If inference begins at step $t$, then the resulting action chunk will not be available until step $t + d$, and so the first $d$ actions cannot actually be executed. However, so long as $d \leq H - s$, these first $d$ timesteps will correspond to actions from the previous chunk that can be executed in the meantime. We call these $d$ actions from the previous chunk that overlap with the current chunk the \textit{action prefix} (see Figure~\ref{fig:diagram}).

We consider policies trained with conditional flow matching \citep{lipman2022flow}, which minimizes the following loss:
\begin{align}
  \bA_t^\tau &= \tau \bA_t + (1 - \tau) \beps & \beps \sim \mathcal{N}(\mathbf{0}, \bI) \\
  \mathcal{L}(\theta) &= \E \; ||\bv_\theta (\bA_t^\tau, \bo_t, \tau) - (\beps - \bA_t)||^2
\end{align}
where $\bv_\theta$ is a neural network and $\tau$ denotes the flow matching timestep. At inference time, $\bv_\theta$ can be integrated from $\tau = 0$ to 1 to produce samples from the dataset distribution $p(\bA_t | \bo_t)$.

\begin{figure}[tbp]
  \centering
  \includegraphics[width=0.46\textwidth]{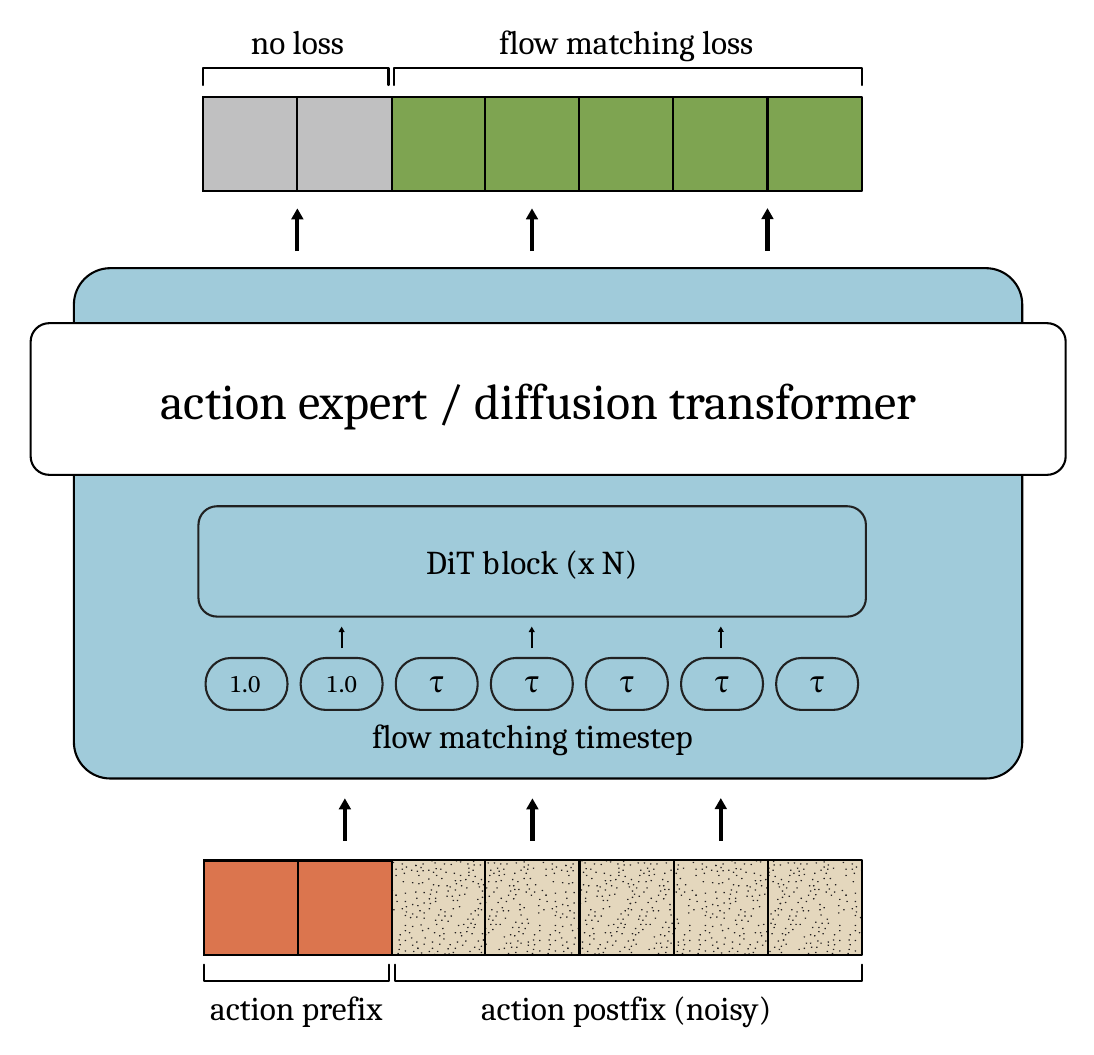}
  \caption{An illustration of our conditioning architecture, as applied to a standard diffusion transformer such as the \PiZeroSix{} action expert. We always feed in ground-truth, non-noisy prefix actions, while learning to denoise the postfix actions. The flow matching timestep differs between tokens, which indicates the inference delay to the model.}
  \label{fig:architecture}
\end{figure}

\section{Training-Time Action Conditioning}
Inference-time RTC \citep{black2025real} conditions the policy on the action prefix (Figure~\ref{fig:diagram}, red) using a inference-time inpainting method based on pseudoinverse guidance \citep{pokle2023training,song2023pseudoinverse}. For improved continuity between chunks, inference-time RTC additionally conditions on \textit{all} overlapping actions, using exponentially decreasing weights for actions beyond the prefix (Figure~\ref{fig:diagram}, yellow). In RTC, this is referred to as ``soft masking''. While pseudoinverse guidance affords great flexibility --- enabling soft masking --- it also requires computing a vector-Jacobian product (using backpropagation) during each denoising step.

The core insight of this work is that we can condition the policy on action prefixes at training time by simulating inference delay. While this does not afford the same flexibility as inference-time inpainting, it eliminates the computational overhead. Formally, we can learn $p(\bA_{t+d:H} | \bo_t, \bA_{t:t+d})$, where $\bA_{t:t+d}$ is an action prefix (Figure~\ref{fig:diagram}, red) and $\bA_{t+d:H}$ is an action postfix (Figure~\ref{fig:diagram}, yellow and green), both taken from the same ground-truth action chunk. Implementing this for most standard policy architectures only requires 3 minimal changes:
\begin{enumerate}
  \item Modify the model architecture to allow for a different flow matching timestep for each action timestep. For a diffusion-transformer-like architecture \citep{peebles2023scalable}, which uses adaLN-zero conditioning for the flow matching timestep, this is trivial --- simply allow the scale, shift, and gate to differ between tokens. This does not change the number of learnable parameters.
  \item Use ground-truth, non-noisy actions for the prefix, and set the corresponding flow matching timesteps to 1. Do not change anything for the postfix. This conditions the model on the ground-truth action prefix while using it to denoise only the postfix.
  \item Mask the loss function so that loss is only computed on outputs corresponding to the postfix.
\end{enumerate}

See Figure~\ref{fig:architecture} for an illustration of this conditioning scheme as applied to a standard diffusion-transformer-like achitecture (e.g., the \PiZeroSix{} action expert). See Algorithm~\ref{alg:training-time-rtc} for Python code fully implementing loss calculation and action generation. In practice, since we do not know the exact inference delay ahead of time (and inference delays in the real world may vary), we sample $d$ randomly during training.

With these modifications, action generation takes as input an action prefix $\bA_{t:t+d}$ and the delay itself $d$ and produces as output an action postfix $\bA_{t+d:H}$. As such, it adheres to the same interface as the action generation component of inference-time RTC (see \citep{black2025real}, Algorithm 1) and thus acts as a seamless drop-in replacement.

\section{Experiments}
In our experiments, we aim to compare training-time RTC to inference-time RTC, as well as to naive synchronous and asynchronous baselines. Our simulated experiments use the same dynamic Kinetix \citep{matthews2024kinetix} benchmark as RTC (see \citep{black2025real} for details). Our real-world experiments build on the \PiZeroSix{} base model \cite{pi06}, and include two precise and challenging tasks: box building and espresso making. We use the same experimental setup as \PiStarZeroSix{} \citep{amin2025pi}.

\begin{figure}[thbp]
  \centering
  \includegraphics[width=0.4\textwidth]{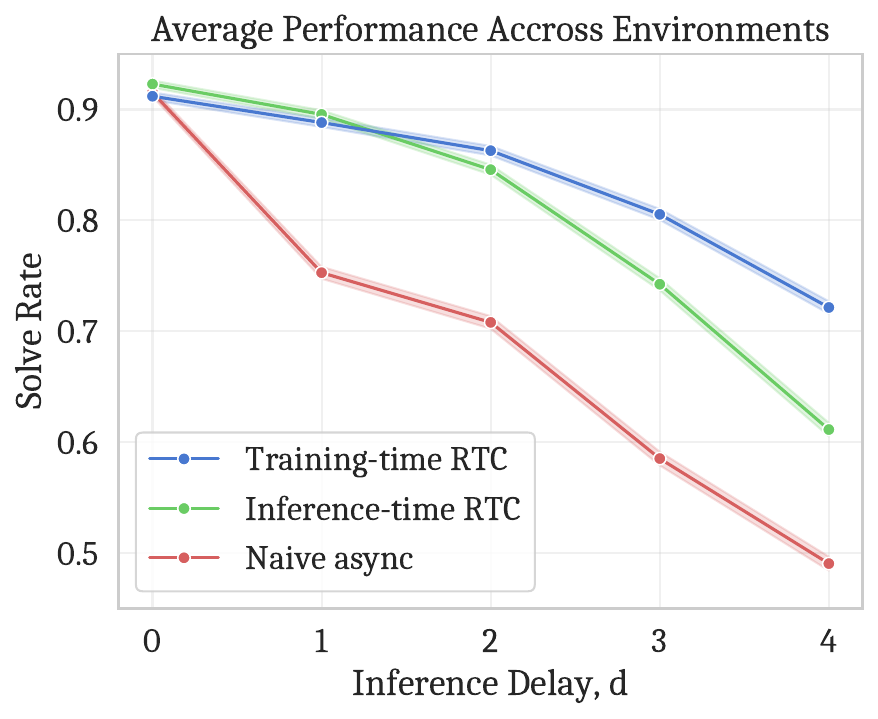}
  \caption{Simulated results: inference delay vs. solve rate with a fixed execution horizon of $s = \max(d, 1)$. Training-time RTC performs better than inference-time RTC at inference delays of 2 or higher.
  Each data point represents 2048 trials, and 95\% Wilson score intervals are shaded in.}
  \label{fig:sim_results}
\end{figure}

\subsection{Simulated Results}
In the dynamic Kinetix benchmark, following RTC \citep{black2025real}, we train action chunking flow policies with a prediction horizon of $H = 8$ and a 4-layer MLP-Mixer \citep{tolstikhin2021mlp} architecture for 32 epochs on data generated by a mixture of expert policies. We report binary success rates with 2048 rollouts per data point and test delays between 0 (fully closed-loop) and 4 (the maximum supported when $H = 8$). Naive asynchronous and inference-time RTC both use the same checkpoint, which is trained normally \textit{without} action prefix conditioning for 32 epochs.

For training-time RTC, we resume training from the 24th epoch and fine-tune for 8 epochs with action prefix conditioning. We do this so that all methods are matched in training compute. We sample delays from $\{0, 1, 2, 3, 4\}$ with exponentially decreasing weights, as we found that higher delays need less training supervision. Better results could likely be obtained by spending more training compute training individual checkpoints for each delay.

The results are presented in Figure~\ref{fig:sim_results}. We find that training-time RTC outperforms inference-time RTC at inference delays of 2 and higher --- with the gap significantly widening as the delay increases. This is likely because, as the size of the prefix grows, the inpainting algorithm has to ``work harder'' to produce a consistent postfix. In these cases, the training-time algorithm is more robust than the pure inference-time algorithm, which relies on a linearization obtained from the Jacobian of the model.
Training-time RTC performs very marginally worse at delays of 1 and 0, likely because training-time RTC does not always receive training supervision for every action --- i.e., slightly less training compute is spent learning to generate the first and second actions.

\begin{figure}[thbp]
  \centering
  \begin{subfigure}{0.48\columnwidth}
    \centering
    \includegraphics[width=\linewidth]{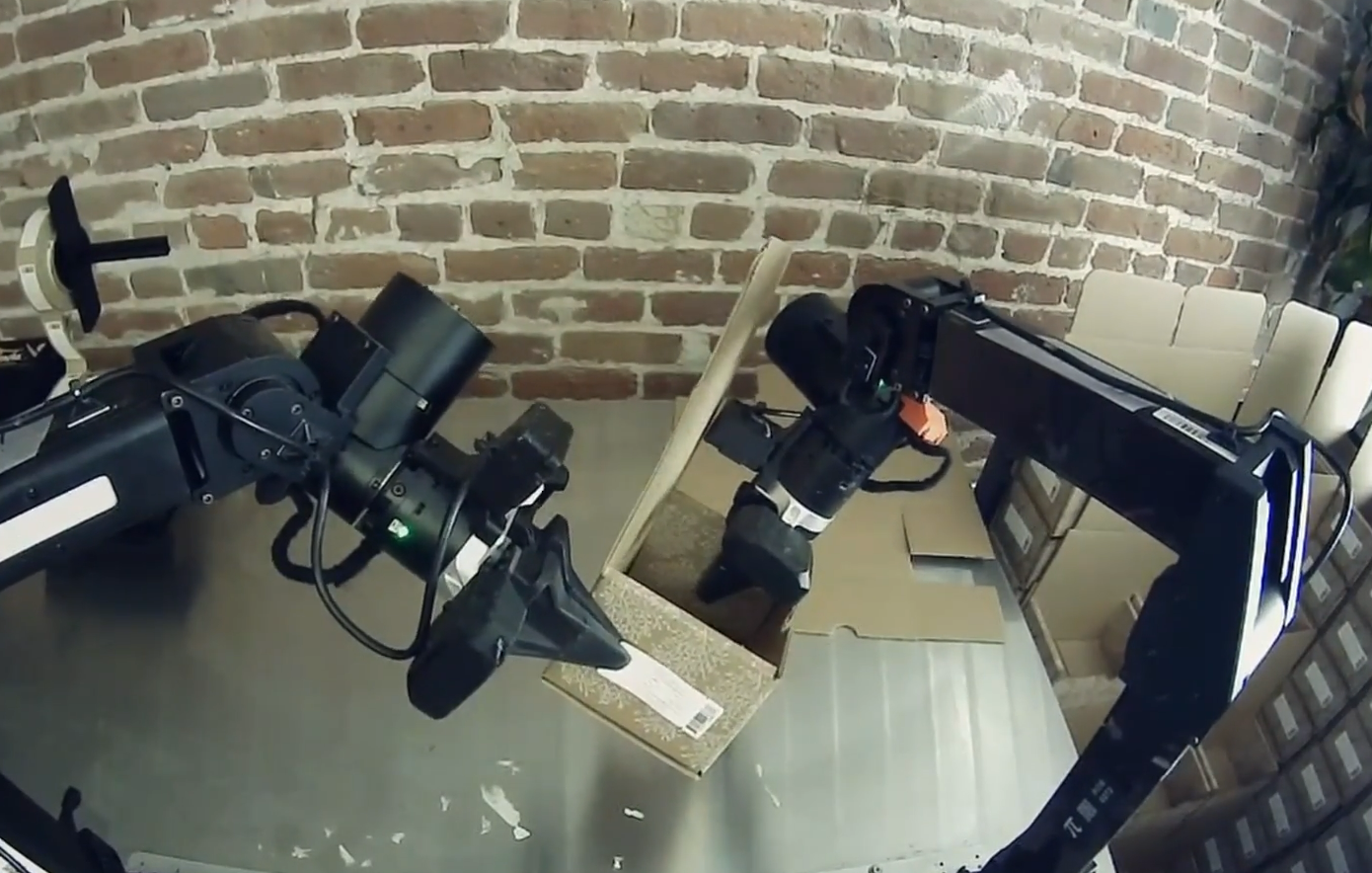}
    \caption{Box building task}
    \label{fig:box_task}
  \end{subfigure}
  \hfill
  \begin{subfigure}{0.48\columnwidth}
    \centering
    \includegraphics[width=\linewidth]{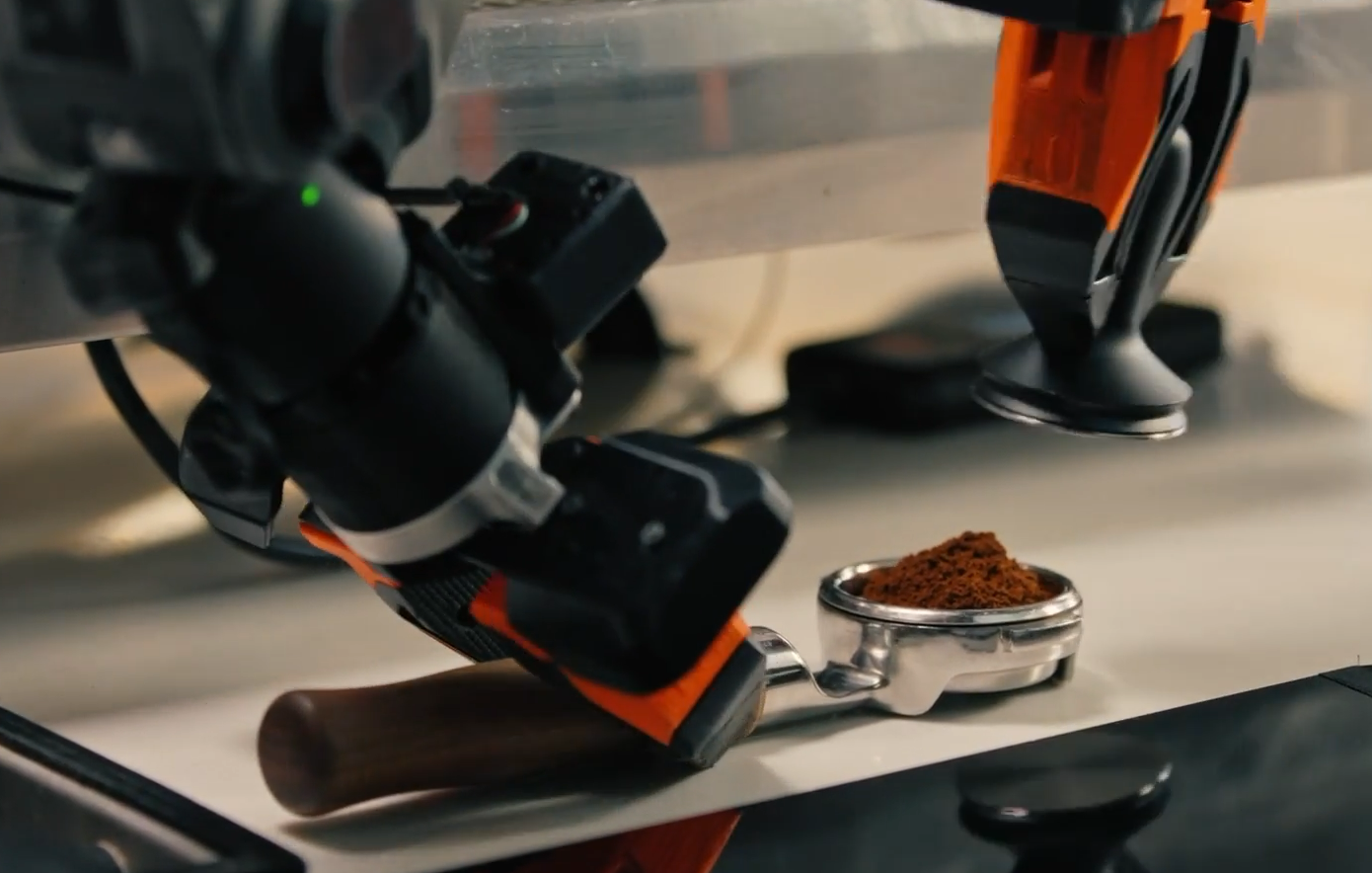}
    \caption{Espresso making task}
    \label{fig:coffee_task}
  \end{subfigure}
  \caption{Real-world evaluation tasks: building a cardboard box and making espresso (including grinding, tamping, extracting, and pouring).}
  \label{fig:real_world_tasks}
\end{figure}

\subsection{Real-World Results}
In our real-world experiments, we use the \PiZeroSix{} base model \cite{pi06} and test on the espresso making and box building tasks from \PiStarZeroSix{} \citep{amin2025pi}; see Figure~\ref{fig:real_world_tasks} for an illustration. As in the simulated experiments, we use the same checkpoint for the synchronous baseline and inference-time RTC, and train a second checkpoint with action prefix conditioning for training-time RTC. Both checkpoints are fine-tuned from the base model on the target task for 8,000 gradient steps with a batch size of 512. We sample delays uniformly between 0 and 10 during training, which supports a maximum latency of 200ms on a 50Hz robot. During evaluations, we perform inference on a remote H100 server with 5 denoising steps, averaging 108ms of end-to-end latency for training-time RTC ($d \approx 5$) and 135ms for inference-time RTC ($d \approx 7$).

\begin{figure}[hp]
  \centering
  \includegraphics[width=0.45\textwidth]{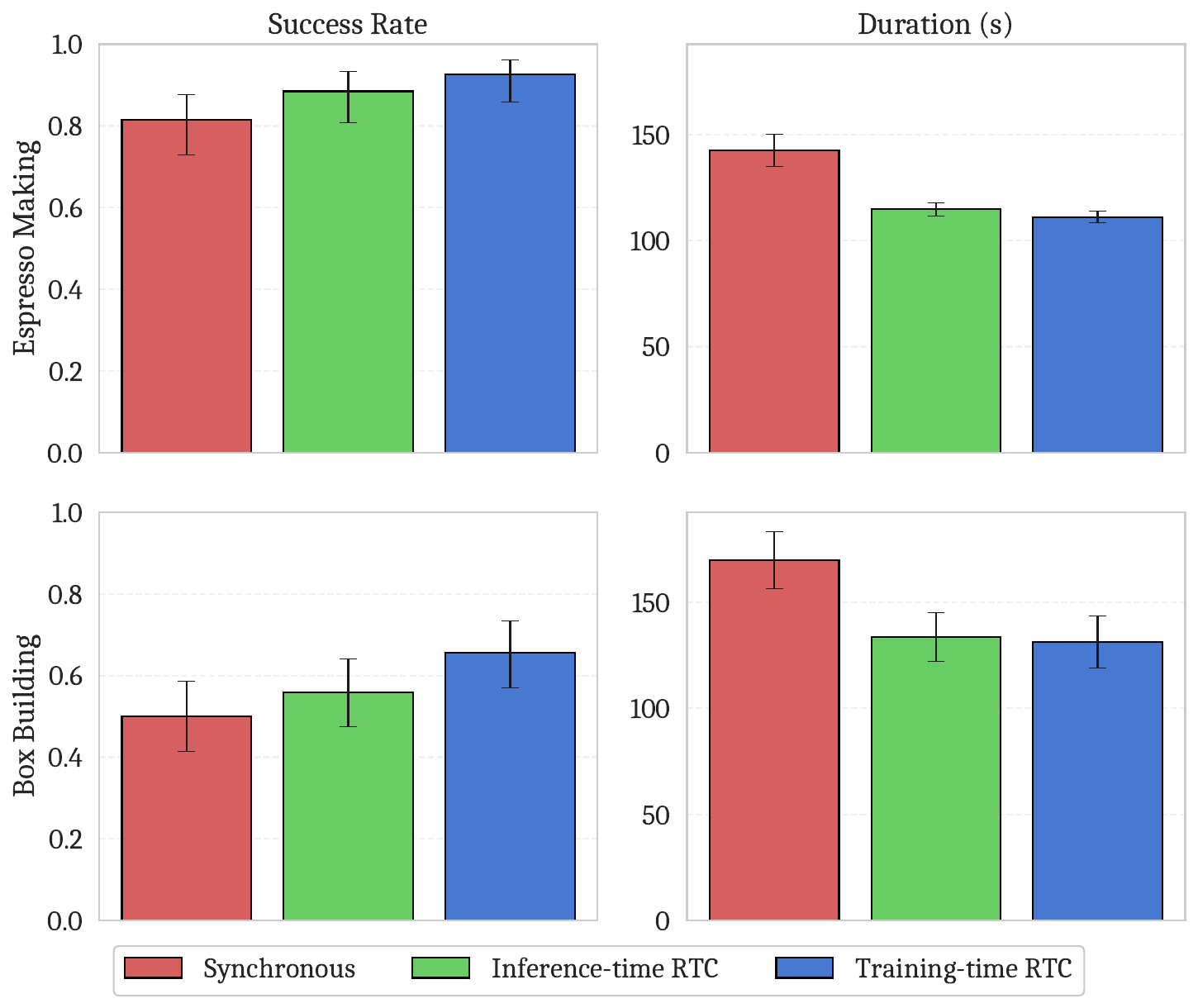}
  \caption{Real-world results: success rate and duration for espresso making and box building. Training-time and inference-time RTC perform similarly, while both improving speed over synchronous inference. Error bars represent 68\% Wilson score intervals for success rate and $\pm1$ SEM for duration.}
  \label{fig:real_results}
\end{figure}

The results are presented in Figure~\ref{fig:real_results}. We find that training-time RTC maintains both performance and speed parity with inference-time RTC without any computational overhead. Both variants of RTC clearly improve speed over the synchronous inference baseline, which exhibits visible pauses in between chunks.

\section{Discussion and Future Work}
In this work, we have presented a simple and effective drop-in replacement for real-time chunking (RTC) that elides any inference-time computational overhead by adding a small amount of additional training compute. Our method requires no modifications to the model architecture or the robot runtime, and can be implemented with only a few additional lines of code. Our simulated experiments show that training-time RTC outperforms inference-time RTC at higher inference delays, while our real-world experiments show that training-time RTC maintains both performance and speed parity with inference-time RTC without any computational overhead.

However, training-time RTC is fundamentally less flexible than inference-time RTC; it only supports conditioning on a ``hard'' action prefix corresponding to the inference delay, whereas inference-time RTC can ``softly'' incorporate additional actions beyond the prefix. Additionally, training-time RTC requires carefully choosing the distribution of delays to simulate at training time based on the expected inference latency. We look forward to future work that can address these limitations and incorporate the best of both worlds.

\section{Acknowledgments}
We thank Laura Smith for developing the espresso making tasks, and helping with early evaluations. We thank Brian Ichter for feedback on the manuscript. As always, we thank our entire team of robot operators for their contributions to data collection and evaluations.

\bibliographystyle{plainnat}
\bibliography{references}

\begin{thebibliography}{29}
\providecommand{\natexlab}[1]{#1}
\providecommand{\url}[1]{\texttt{#1}}
\expandafter\ifx\csname urlstyle\endcsname\relax
  \providecommand{\doi}[1]{doi: #1}\else
  \providecommand{\doi}{doi: \begingroup \urlstyle{rm}\Url}\fi

\bibitem[Amin et~al.(2025)Amin, Aniceto, Balakrishna, Black, Conley, Connors, Darpinian, Dhabalia, DiCarlo, Driess, et~al.]{amin2025pi}
Ali Amin, Raichelle Aniceto, Ashwin Balakrishna, Kevin Black, Ken Conley, Grace Connors, James Darpinian, Karan Dhabalia, Jared DiCarlo, Danny Driess, et~al.
\newblock $\pi^{*}_{0.6}$: a vla that learns from experience.
\newblock \emph{arXiv preprint arXiv:2511.14759}, 2025.

\bibitem[Belkhale and Sadigh(2024)]{belkhale2024minivla}
Suneel Belkhale and Dorsa Sadigh.
\newblock Minivla: A better vla with a smaller footprint, 2024.
\newblock URL \url{https://github.com/Stanford-ILIAD/openvla-mini}.

\bibitem[Bjorck et~al.(2025)Bjorck, Casta{\~n}eda, Cherniadev, Da, Ding, Fan, Fang, Fox, Hu, Huang, et~al.]{bjorck2025gr00t}
Johan Bjorck, Fernando Casta{\~n}eda, Nikita Cherniadev, Xingye Da, Runyu Ding, Linxi Fan, Yu~Fang, Dieter Fox, Fengyuan Hu, Spencer Huang, et~al.
\newblock Gr00t n1: An open foundation model for generalist humanoid robots.
\newblock \emph{arXiv preprint arXiv:2503.14734}, 2025.

\bibitem[Black et~al.(2024)Black, Brown, Driess, Esmail, Equi, Finn, Fusai, Groom, Hausman, Ichter, et~al.]{black2024pi}
Kevin Black, Noah Brown, Danny Driess, Adnan Esmail, Michael Equi, Chelsea Finn, Niccolo Fusai, Lachy Groom, Karol Hausman, Brian Ichter, et~al.
\newblock $\pi_0$: A vision-language-action flow model for general robot control.
\newblock \emph{arXiv preprint arXiv:2410.24164}, 2024.

\bibitem[Black et~al.(2025)Black, Galliker, and Levine]{black2025real}
Kevin Black, Manuel~Y Galliker, and Sergey Levine.
\newblock Real-time execution of action chunking flow policies.
\newblock \emph{arXiv preprint arXiv:2506.07339}, 2025.

\bibitem[Brohan et~al.(2023)Brohan, Brown, Carbajal, Chebotar, Chen, Choromanski, Ding, Driess, Dubey, Finn, Florence, Fu, Arenas, Gopalakrishnan, Han, Hausman, Herzog, Hsu, Ichter, Irpan, Joshi, Julian, Kalashnikov, Kuang, Leal, Lee, Lee, Levine, Lu, Michalewski, Mordatch, Pertsch, Rao, Reymann, Ryoo, Salazar, Sanketi, Sermanet, Singh, Singh, Soricut, Tran, Vanhoucke, Vuong, Wahid, Welker, Wohlhart, Wu, Xia, Xiao, Xu, Xu, Yu, and Zitkovich]{rt22023arxiv}
Anthony Brohan, Noah Brown, Justice Carbajal, Yevgen Chebotar, Xi~Chen, Krzysztof Choromanski, Tianli Ding, Danny Driess, Avinava Dubey, Chelsea Finn, Pete Florence, Chuyuan Fu, Montse~Gonzalez Arenas, Keerthana Gopalakrishnan, Kehang Han, Karol Hausman, Alex Herzog, Jasmine Hsu, Brian Ichter, Alex Irpan, Nikhil Joshi, Ryan Julian, Dmitry Kalashnikov, Yuheng Kuang, Isabel Leal, Lisa Lee, Tsang-Wei~Edward Lee, Sergey Levine, Yao Lu, Henryk Michalewski, Igor Mordatch, Karl Pertsch, Kanishka Rao, Krista Reymann, Michael Ryoo, Grecia Salazar, Pannag Sanketi, Pierre Sermanet, Jaspiar Singh, Anikait Singh, Radu Soricut, Huong Tran, Vincent Vanhoucke, Quan Vuong, Ayzaan Wahid, Stefan Welker, Paul Wohlhart, Jialin Wu, Fei Xia, Ted Xiao, Peng Xu, Sichun Xu, Tianhe Yu, and Brianna Zitkovich.
\newblock Rt-2: Vision-language-action models transfer web knowledge to robotic control.
\newblock In \emph{arXiv preprint arXiv:2307.15818}, 2023.

\bibitem[Cheang et~al.(2024)Cheang, Chen, Jing, Kong, Li, Li, Liu, Wu, Xu, Yang, Zhang, and Zhu]{cheang2024gr2generativevideolanguageactionmodel}
Chi-Lam Cheang, Guangzeng Chen, Ya~Jing, Tao Kong, Hang Li, Yifeng Li, Yuxiao Liu, Hongtao Wu, Jiafeng Xu, Yichu Yang, Hanbo Zhang, and Minzhao Zhu.
\newblock Gr-2: A generative video-language-action model with web-scale knowledge for robot manipulation.
\newblock \emph{arXiv preprint arXiv:2410.06158}, 2024.

\bibitem[Cheng et~al.(2024)Cheng, Ji, Yang, Zou, Kautz, Biyik, Yin, Liu, and Wang]{cheng2024navila}
An-Chieh Cheng, Yandong Ji, Zhaojing Yang, Xueyan Zou, Jan Kautz, Erdem Biyik, Hongxu Yin, Sifei Liu, and Xiaolong Wang.
\newblock {NaVILA: Legged Robot Vision-Language-Action Model for Navigation}.
\newblock \emph{arXiv preprint arXiv:2412.04453}, 2024.

\bibitem[Chi et~al.(2023)Chi, Xu, Feng, Cousineau, Du, Burchfiel, Tedrake, and Song]{chi2023diffusion}
Cheng Chi, Zhenjia Xu, Siyuan Feng, Eric Cousineau, Yilun Du, Benjamin Burchfiel, Russ Tedrake, and Shuran Song.
\newblock Diffusion policy: Visuomotor policy learning via action diffusion.
\newblock \emph{The International Journal of Robotics Research}, page 02783649241273668, 2023.

\bibitem[Collaboration et~al.(2023)Collaboration, Padalkar, Pooley, Jain, Bewley, Herzog, Irpan, Khazatsky, Rai, Singh, et~al.]{collaboration2023open}
OX-Embodiment Collaboration, A~Padalkar, A~Pooley, A~Jain, A~Bewley, A~Herzog, A~Irpan, A~Khazatsky, A~Rai, A~Singh, et~al.
\newblock {Open X-Embodiment}: Robotic learning datasets and {RT-X} models.
\newblock \emph{arXiv preprint arXiv:2310.08864}, 1\penalty0 (2), 2023.

\bibitem[Intelligence et~al.(2025)Intelligence, Black, Brown, Darpinian, Dhabalia, Driess, Esmail, Equi, Finn, Fusai, et~al.]{intelligence2025pi}
Physical Intelligence, Kevin Black, Noah Brown, James Darpinian, Karan Dhabalia, Danny Driess, Adnan Esmail, Michael Equi, Chelsea Finn, Niccolo Fusai, et~al.
\newblock $\pi_{0.5}$: A vision-language-action model with open-world generalization.
\newblock \emph{arXiv preprint arXiv:2504.16054}, 2025.

\bibitem[Kim et~al.(2024)Kim, Pertsch, Karamcheti, Xiao, Balakrishna, Nair, Rafailov, Foster, Lam, Sanketi, et~al.]{kim2024openvla}
Moo~Jin Kim, Karl Pertsch, Siddharth Karamcheti, Ted Xiao, Ashwin Balakrishna, Suraj Nair, Rafael Rafailov, Ethan Foster, Grace Lam, Pannag Sanketi, et~al.
\newblock Openvla: An open-source vision-language-action model.
\newblock \emph{arXiv preprint arXiv:2406.09246}, 2024.

\bibitem[Lipman et~al.(2022)Lipman, Chen, Ben-Hamu, Nickel, and Le]{lipman2022flow}
Yaron Lipman, Ricky~TQ Chen, Heli Ben-Hamu, Maximilian Nickel, and Matt Le.
\newblock Flow matching for generative modeling.
\newblock \emph{arXiv preprint arXiv:2210.02747}, 2022.

\bibitem[Liu et~al.(2024)Liu, Wu, Li, Tan, Chen, Wang, Xu, Su, and Zhu]{liu2024rdt}
Songming Liu, Lingxuan Wu, Bangguo Li, Hengkai Tan, Huayu Chen, Zhengyi Wang, Ke~Xu, Hang Su, and Jun Zhu.
\newblock Rdt-1b: a diffusion foundation model for bimanual manipulation.
\newblock \emph{arXiv preprint arXiv:2410.07864}, 2024.

\bibitem[Matthews et~al.(2024)Matthews, Beukman, Lu, and Foerster]{matthews2024kinetix}
Michael Matthews, Michael Beukman, Chris Lu, and Jakob Foerster.
\newblock Kinetix: Investigating the training of general agents through open-ended physics-based control tasks.
\newblock \emph{arXiv preprint arXiv:2410.23208}, 2024.

\bibitem[Peebles and Xie(2023)]{peebles2023scalable}
William Peebles and Saining Xie.
\newblock Scalable diffusion models with transformers.
\newblock In \emph{Proceedings of the IEEE/CVF international conference on computer vision}, pages 4195--4205, 2023.

\bibitem[Pertsch et~al.(2025)Pertsch, Stachowicz, Ichter, Driess, Nair, Vuong, Mees, Finn, and Levine]{pertsch2025fast}
Karl Pertsch, Kyle Stachowicz, Brian Ichter, Danny Driess, Suraj Nair, Quan Vuong, Oier Mees, Chelsea Finn, and Sergey Levine.
\newblock Fast: Efficient action tokenization for vision-language-action models.
\newblock \emph{arXiv preprint arXiv:2501.09747}, 2025.

\bibitem[Pokle et~al.(2023)Pokle, Muckley, Chen, and Karrer]{pokle2023training}
Ashwini Pokle, Matthew~J Muckley, Ricky~TQ Chen, and Brian Karrer.
\newblock Training-free linear image inverses via flows.
\newblock \emph{arXiv preprint arXiv:2310.04432}, 2023.

\bibitem[Sendai et~al.(2025)Sendai, Alvarez, Matsushima, Matsuo, and Iwasawa]{sendai2025leave}
Kohei Sendai, Maxime Alvarez, Tatsuya Matsushima, Yutaka Matsuo, and Yusuke Iwasawa.
\newblock Leave no observation behind: Real-time correction for vla action chunks.
\newblock \emph{arXiv preprint arXiv:2509.23224}, 2025.

\bibitem[Shukor et~al.(2025)Shukor, Aubakirova, Capuano, Kooijmans, Palma, Zouitine, Aractingi, Pascal, Russi, Marafioti, et~al.]{shukor2025smolvla}
Mustafa Shukor, Dana Aubakirova, Francesco Capuano, Pepijn Kooijmans, Steven Palma, Adil Zouitine, Michel Aractingi, Caroline Pascal, Martino Russi, Andres Marafioti, et~al.
\newblock Smolvla: A vision-language-action model for affordable and efficient robotics.
\newblock \emph{arXiv preprint arXiv:2506.01844}, 2025.

\bibitem[Song et~al.(2023)Song, Vahdat, Mardani, and Kautz]{song2023pseudoinverse}
Jiaming Song, Arash Vahdat, Morteza Mardani, and Jan Kautz.
\newblock Pseudoinverse-guided diffusion models for inverse problems.
\newblock In \emph{International Conference on Learning Representations}, 2023.

\bibitem[Tang et~al.(2025)Tang, Sun, Zhao, Yang, Lin, Zhang, Hou, Lu, Liu, and Han]{tang2025vlash}
Jiaming Tang, Yufei Sun, Yilong Zhao, Shang Yang, Yujun Lin, Zhuoyang Zhang, James Hou, Yao Lu, Zhijian Liu, and Song Han.
\newblock Vlash: Real-time vlas via future-state-aware asynchronous inference.
\newblock \emph{arXiv preprint arXiv:2512.01031}, 2025.

\bibitem[Team et~al.(2025)Team, Abeyruwan, Ainslie, Alayrac, Arenas, Armstrong, Balakrishna, Baruch, Bauza, Blokzijl, et~al.]{team2025gemini}
Gemini~Robotics Team, Saminda Abeyruwan, Joshua Ainslie, Jean-Baptiste Alayrac, Montserrat~Gonzalez Arenas, Travis Armstrong, Ashwin Balakrishna, Robert Baruch, Maria Bauza, Michiel Blokzijl, et~al.
\newblock Gemini robotics: Bringing ai into the physical world.
\newblock \emph{arXiv preprint arXiv:2503.20020}, 2025.

\bibitem[team(2025)]{pi06}
Physical~Intelligence team.
\newblock $\pi_{0.6}$ model card.
\newblock 2025.

\bibitem[Tolstikhin et~al.(2021)Tolstikhin, Houlsby, Kolesnikov, Beyer, Zhai, Unterthiner, Yung, Steiner, Keysers, Uszkoreit, et~al.]{tolstikhin2021mlp}
Ilya~O Tolstikhin, Neil Houlsby, Alexander Kolesnikov, Lucas Beyer, Xiaohua Zhai, Thomas Unterthiner, Jessica Yung, Andreas Steiner, Daniel Keysers, Jakob Uszkoreit, et~al.
\newblock Mlp-mixer: An all-mlp architecture for vision.
\newblock \emph{Advances in neural information processing systems}, 34:\penalty0 24261--24272, 2021.

\bibitem[Zhao et~al.(2023)Zhao, Kumar, Levine, and Finn]{zhao2023learning}
Tony~Z Zhao, Vikash Kumar, Sergey Levine, and Chelsea Finn.
\newblock Learning fine-grained bimanual manipulation with low-cost hardware.
\newblock \emph{arXiv preprint arXiv:2304.13705}, 2023.

\bibitem[Zhao et~al.(2024)Zhao, Tompson, Driess, Florence, Ghasemipour, Finn, and Wahid]{zhao2024aloha}
Tony~Z Zhao, Jonathan Tompson, Danny Driess, Pete Florence, Kamyar Ghasemipour, Chelsea Finn, and Ayzaan Wahid.
\newblock Aloha unleashed: A simple recipe for robot dexterity.
\newblock \emph{arXiv preprint arXiv:2410.13126}, 2024.

\bibitem[Zhen et~al.(2024)Zhen, Qiu, Chen, Yang, Yan, Du, Hong, and Gan]{zhen20243dvla}
Haoyu Zhen, Xiaowen Qiu, Peihao Chen, Jincheng Yang, Xin Yan, Yilun Du, Yining Hong, and Chuang Gan.
\newblock 3d-vla: 3d vision-language-action generative world model.
\newblock \emph{arXiv preprint arXiv:2403.09631}, 2024.

\bibitem[Zheng et~al.(2024)Zheng, Liang, Huang, Gao, Daum{\'e}~III, Kolobov, Huang, and Yang]{zheng2024tracevla}
Ruijie Zheng, Yongyuan Liang, Shuaiyi Huang, Jianfeng Gao, Hal Daum{\'e}~III, Andrey Kolobov, Furong Huang, and Jianwei Yang.
\newblock Tracevla: Visual trace prompting enhances spatial-temporal awareness for generalist robotic policies.
\newblock \emph{arXiv preprint arXiv:2412.10345}, 2024.

\end{thebibliography}

\begin{algorithm*}[htbp]
\caption{Python code implementing the loss and sampling functions for training-time action conditioning. Differences from standard flow matching code are highlighted in red.}
\label{alg:training-time-rtc}
\begin{lstlisting}
import jax
import jax.numpy as jnp

def compute_loss(rng, model, observation, action_chunk, max_delay):
  b, ah, ad = action_chunk.shape # (batch_size, action_horizon, action_dim)
  noise_rng, time_rng, delay_rng = jax.random.split(rng)
  time = jax.random.uniform(time_rng, (b,))
  noise = jax.random.normal(noise_rng, (b, ah, ad))
  # sample delays from some distribution of choice:
  # here, we use Unif[0, max_delay), as in our real-world experiments
  (*@\textcolor{red}{\ttfamily delay = jax.random.randint(delay\_rng, (b,), 0, max\_delay)}@*)

  # set time to 1.0 for the action prefix
  # time becomes shape (batch_size, action_horizon)
  (*@\textcolor{red}{\ttfamily prefix\_mask = jnp.arange(ah)[None, :] < delay[:, None]}@*)
  (*@\textcolor{red}{\ttfamily time = jnp.where(prefix\_mask, 1.0, time[:, None])}@*)

  # compute the noisy action postfix and run the model
  x_t = time[:, :, None] * action_chunk + (1 - time[:, :, None]) * noise
  pred_v_t = model(observation, x_t, time)
  loss = (pred_v_t - (action_chunk - noise))**2

  # compute the loss on the postfix only
  (*@\textcolor{red}{\ttfamily postfix\_mask = jnp.logical\_not(prefix\_mask)[:, :, None]}@*)
  (*@\textcolor{red}{\ttfamily loss = jnp.sum(loss * postfix\_mask) / (jnp.sum(postfix\_mask) + 1e-8)}@*)
  return loss

def sample_actions(rng, model, observation, action_prefix, delay, num_steps):
  # assume action_prefix is padded to (batch_size, action_horizon, action_dim),
  # but only the first delay actions are valid
  b, ah, ad = action_prefix.shape
  x_t = jax.random.normal(rng, (b, ah, ad))
  time = 0.0
  dt = 1 / num_steps
  (*@\textcolor{red}{\ttfamily prefix\_mask = jnp.arange(ah)[None, :] < delay}@*)

  for _ in range(num_steps):
    (*@\textcolor{red}{\ttfamily x\_t = jnp.where(prefix\_mask[:, :, None], action\_prefix, x\_t)}@*)
    (*@\textcolor{red}{\ttfamily time\_masked = jnp.where(prefix\_mask, 1.0, time)}@*)
    v_t = model(observation, x_t, time_masked)
    x_t = x_t + dt * v_t
    time = time + dt

  return x_t
\end{lstlisting}
\end{algorithm*}

\end{document}